# Advanced Machine Learning Techniques for Social Support Detection on Social Media

Olga Kolesnikova[a,*], Moein Shahiki Tash[a,*], Zahra Ahani[a], Ameeta Agrawal[b], Raúl Monroy[c], Grigori Sidorov[a]

[a]*Instituto Politécnico Nacional (IPN), Centro de Investigación en Computación (CIC), Mexico City, Mexico,*
[b]*Department of Computer Science, Portland State University, USA,*
[c]*Tecnologico de Monterrey Escuela de Ingenieria y Ciencias*

**Abstract**

The widespread use of social media highlights the need to understand its impact, particularly the role of online social support. This study uses a dataset focused on online social support, which includes binary and multiclass classifications of social support content on social media. The classification of social support is divided into three tasks. The first task focuses on distinguishing between supportive and non-supportive. The second task aims to identify whether the support is directed toward an individual or a group. The third task categorizes the specific type of social support, grouping it into categories such as Nation, LGBTQ, Black people, Women, Religion, and Other (if it does not fit into the previously mentioned categories). To address data imbalances in these tasks, we employed K-means clustering for balancing the dataset and compared the results with the original unbalanced data. Using advanced machine learning techniques, including transformers and zero-shot learning approaches with GPT3, GPT4, and GPT4-o, we predict social support levels in various contexts. The effectiveness of the dataset is evaluated using baseline models across different learning approaches, with transformer-based methods demonstrating superior performance. Additionally, we achieved a 0.4% increase in the macro F1 score for the second task and a 0.7% increase for the third task, compared to previous work utilizing traditional machine learning with psycholinguistic and unigram-based TF-IDF values.

## 1. Introduction

Platforms such as YouTube, Instagram, Snapchat, Tik-Tok, and Facebook have become integral to daily life worldwide (Statista, 2024). These platforms allow users to observe and interact with others, obtaining social rewards (Meshi et al., 2015) that act as reinforcers, bringing people back to these sites repeatedly and for substantial durations (Stewart, 2016). However, problematic social media use has been consistently linked to negative mental health outcomes. A meta-analysis concluded that greater problematic Facebook use is associated with higher levels of depression and anxiety (Marino et al., 2018). Another recent meta-analysis, which focused on problematic use across any social media platform, found similar associations, indicating that greater problematic social media use correlates with increased depression, anxiety, and loneliness (Huang, 2022).

One potential mechanism for mediating this relationship could be the amount of social support an individual receives. Researchers have speculated that more time on social media leads to a lack of face-to-face social interaction and that this lack of in-person social support may be responsible for the association between problematic social media use and negative mental health (Shensa et al., 2017). Indeed, more time spent on social media has been associated with less perceived real-life social support (McDougall et al., 2016).

Despite these findings, more comprehensive data and an understanding of the complex relationship between social media use and mental health is still needed. To address this gap (Ahani et al., 2024b) have introduced a new dataset focused on online social support. Online social support involves behaviors, communication, and interactions that demonstrate care and appreciation for individuals, thus fostering a sense of belonging and helping to cope with life's challenges (Ko et al., 2013). This dataset features two binary levels: the *first* distinguishes between social support and non-social support, while the *second* categorizes social support into individual and group support. The group support category is further divided into six subcategories: Nation, Women, Black People, LGBTQ, Religion, and Other. In their study Ahani et al. (2024b) applied a range of machine learning techniques to classify the data, including traditional classifiers.

This research employs advanced techniques such as Transformers, and zero-shot learning approaches with models like GPT-3, GPT-4, and GPT-4-o. The aim is to develop prediction models for social support across various contexts.

The Transformer model was initially applied to natural language processing (NLP) tasks, where it delivered significant performance improvements (Devlin et al., 2018). For instance, Vaswani et al. (2017) introduced the Transformer architecture, which is based on the attention mechanism, to tackle machine translation and English constituency parsing tasks. Devlin et al. (2018) later developed BERT (Bidirectional Encoder Representations from Transformers), a new language representation model that pre-trains a transformer on the unlabeled text while considering the context of each word bidirectionally. Upon its release, BERT set new performance benchmarks across 11 NLP tasks. Subsequently, Brown et al. (2020) pre-trained GPT-3 (Generative Pre-trained Transformer 3), a massive transformer-based model with 175 billion parameters, on 45 terabytes of compressed plain text data. GPT-3 demonstrated robust performance across various natural language tasks without the need for fine-tuning. These transformer-based models have made notable advancements in NLP due to their strong representation capabilities. GPT-4, a large multimodal model, is designed to process both image and text inputs and generate text output. Such models are considered a significant area of study due to their potential applications in dialogue systems, text summarization, and machine translation. Consequently, substantial interest and progress have been observed in this field in recent years (Achiam et al., 2023). After conducting tests and experiments, we achieved favorable outcomes across all three tasks using Transformer models. Additionally, we obtained satisfactory results through zero-shot learning with both Transformer-based and GPT-based models. also in this part you can find the Overview of Contributions.

---

*These authors contributed equally to this work.

- Achieves a notable increase in macro F1-scores across all three tasks through the implementation of an advanced machine learning model, demonstrating significant performance improvements over previous methods.

- Using K-means clustering for dataset balancing in multi-label tasks.

- Detailed analysis of class imbalances and their impact on model performance.

**2. Related work**

Social support is the perception that one is cared for, valued, and part of a network of reciprocal obligations (Cobb, 1976).

Ahani et al. (2024b) conducted a study on online social support by collecting data from YouTube comments. The research analyzed comments from 17 videos across various categories such as nation, black people, women, religion, LGBTQ, and others. Initially, 66,272 comments were gathered. After removing duplicates and non-English comments, the dataset was refined to 42,695 comments. From this dataset, 5,000 comments containing specific keywords and another 5,000 randomly selected comments were chosen. Comments associated with the selected videos underwent no additional filtering or selection, allowing for an accurate assessment of the distribution of supportive comments for each video and the exploration of related aspects. The study examined the psycholinguistic, emotional, and sentiment features through Tash et al. (2024b,c): 1. Supportive detection with LIWC 2. Supportive detection with TF-IDF 3. Supportive detection with TF-IDF and LIWC combined. Subsequently, traditional machine learning algorithms were applied. The best results are presented in Table 1.

Hope is an extraordinary human ability that allows individuals to envision future events and their potential outcomes with flexibility. Balouchzahi et al. (2023); Arif et al. (2024) introduced PolyHope, the first multiclass hope speech detection dataset in English. The dataset was developed by collecting around 100,000 English tweets, which were then preprocessed down to approximately 23,000 tweets. From this, a random sample of 10,000 tweets was chosen for annotation. After annotation, the dataset included 4,081 tweets labeled as NotHope, 2,335 as GeneralizedHope, 982 as RealisticHope, and 858 as UnrealisticHope, resulting in a total of 8,256 tweets. To evaluate the dataset's effectiveness, various baseline models were tested using different learning approaches, including traditional machine learning, deep learning Tash et al. (2024a); Ahani et al. (2024a); Ahani et al., and transformer-based methods. The results for the top models in each learning approach, showing the average macro F1 scores for both binary and multiclass classification tasks on the PolyHope dataset, are shown in Table 2.

Zero-shot learning (ZSL) is a difficult task because there is no labeled data available for unseen classes during the training phase. The primary focus of the work by Xiong et al. (2021) is on tackling the Extreme Zero-Shot Multi-label Classification (EZ-XMC) problem, which involves classifying text instances into numerous labels without supervision. The authors introduce MACLR (Multi-scale Adaptive Clustering with Label Regularization), a method that pre-trains Transformer-based encoders using self-supervised contrastive learning to generate effective semantic embeddings. They utilize four public EZ-XMC datasets for their experiments. The MACLR method is compared against several baseline models, including TF-IDF, XR-Linear, GloVe, SentBERT, Paraphrase MPNet, SimCSE, and ICT, as well as few-shot learning models like Astec, SiameseXML, ZestXML, and a fine-tuned SentBERT. MACLR outperformed all other baselines, achieving notable results such as a P@1 score of 18.74% in few-shot scenarios with only 1% sampled labels. The findings indicate that MACLR significantly improves precision and recall across datasets and remains robust even with minimal supervision, demonstrating its effectiveness in domains with many cold-start labels.

Significant advancements in natural language processing have been propelled by large language models (LLMs), sparking widespread interest in artificial intelligence among both academics and the general public since the launch of OpenAI's ChatGPT in late 2022.

The use of LLMs in computational sociology, specifically for supervised text classification tasks, is explored by Chae and Davidson (2023). Four LLM architectures were evaluated: two models based on the BERT framework and two variants of OpenAI's GPT-3, namely Ada and Davinci. Social media posts about politicians were used as data, sourced from a widely used Twitter dataset consisting of 1,691 tweets and a new collection of 2,400 Facebook comments, both centered on the 2016 US Presidential election. Experiments ranged from prompt-based zero-shot learning to fine-tuning with thousands of annotated examples. It was found that LLMs could achieve high accuracy in text classification, often surpassing traditional machine learning baselines. Key findings included the efficiency of fine-tuning smaller models for cost-effective yet accurate performance and the necessity of evaluating model biases and ensuring transparency and reproducibility in research. The best result is achieved by GPT-3 Davinci, GPT-3 Ada, and BART-MNLI

Table 1: The most effective outcomes for categorizing social support

| Model | Task | Feature | Weighted scores | | | Macro scores | | | Accuracy |
|---|---|---|---|---|---|---|---|---|---|
| | | | precision | recall | F1-score | precision | recall | F1-score | |
| **SVM (linear)** | Task1 | TF-IDF and LIWC | 0.8558 | 0.8626 | 0.8557 | 0.8190 | 0.7602 | 0.7830 | 0.8626 |
| **SVM (linear)** | Task2 | TF-IDF and LIWC | 0.8783 | 0.8828 | 0.8792 | 0.8186 | 0.7808 | 0.7969 | 0.8828 |
| **Soft voting** | Task3 | TF-IDF | 0.8280 | 0.8273 | 0.8239 | 0.8074 | 0.7043 | 0.7262 | 0.8273 |

Table 2: Best performing models reported in PolyHope dataset

| Model | Learning approach | Averaged-weighted F1 | Averaged-macro F1 |
|---|---|---|---|
| **Binary Hope Speech Detection** | | | |
| LR | Machine learning | 0.80 | 0.80 |
| BiLSTM and CNN both with FastText | Deep learning | 0.82 | 0.82 |
| **BERT, RoBERTa, and XLNet** | **Transformers** | **0.85** | **0.85** |
| **Multiclass Hope Speech Detection** | | | |
| CatBoost | Machine learning | 0.64 | 0.54 |
| CNN+GloVe | Deep learning | 0.70 | 0.61 |
| **BERT** | **Transformers** | **0.77** | **0.72** |

models on Facebook (MT) data, with an overall F1 score of 0.75.

Tang et al. (2023) propose a Multi-disciplinary Collaboration (MC) framework aimed at enhancing the proficiency and reasoning capabilities of LLMs in the medical domain. The framework utilizes GPT-3.5-Turbo and GPT-4 models from Azure OpenAI Service and leverages a role-playing setting where LLM-based agents engage in collaborative multi-round discussions. Data was collected from MedQA, MedMCQA, PubMedQA, and six medical-related subtasks from MMLU, covering examinations such as the US Medical Licensing Examination and AIIMS and NEET PG entrance exams. Specifically, the study evaluates the framework using a total of 300 examples randomly sampled from each of these nine datasets: MedQA (12,723 questions), MedMCQA (194,000 questions), PubMedQA (1,000 pairs), and six medical subtasks related to MMLU (anatomy, clinical knowledge, college medicine, medical genetics, professional medicine and college biology). The framework operates in a zero-shot setting, with experiments that demonstrate significant improvements over baseline methods in accuracy, achieving an average accuracy of 67.0% when using the entire MC process. Key findings indicate that the multi-agent approach mitigates issues such as hallucinations and domain knowledge retrieval errors, showcasing its potential for robust medical reasoning applications. The best result is achieved by the method "w/ MedAgents + Anal & Summ & Cons", with an accuracy of 67.0%.

This study Lin et al. (2017) addresses the class imbalance problem in datasets by introducing two clustering-based undersampling strategies: using cluster centers and their nearest neighbors to represent the majority class. It evaluated these strategies with classifiers such as C4.5, k-NN, SVM, Naïve Bayes, and Multilayer Perceptron (MLP), along with ensemble methods like AdaBoost. Experiments conducted on 44 small-scale datasets and two large-scale datasets (Breast Cancer and Protein Homology Prediction) demonstrated that the nearest neighbors strategy consistently outperformed state-of-the-art approaches. For small datasets, the single MLP classifier yielded the best performance, while for large datasets, C4.5 ensembles with the proposed approach achieved the highest accuracy. The results highlight the effectiveness of clustering-based undersampling, particularly when combined with ensemble methods, in improving classification performance on imbalanced data.

## 3. Methodology

### 3.1. Dataset

The dataset utilized in this study, as referenced from Ahani et al. (2024b), was derived from YouTube comments extracted from 17 diverse videos covering topics such as nationality, race, gender, religion, and LGBTQ issues, among others. Initially, a pool of 66,272 comments was gathered, which was subsequently refined to 42,695 comments after eliminating duplicates and non-English comments. Following this, a subset of 10,000 comments containing specific keywords was randomly selected. It's noteworthy that no further filtering or selection processes were applied to comments associated with the chosen videos. This methodology facilitated an accurate examination of the distribution of supportive comments across each video while also exploring associated facets. Additional dataset details are included in Table 4. You can also find more examples across various tasks in the Table 4.

### 3.2. Preprocessing

At first, data preprocessing consisted of eliminating duplicate comments and choosing only English tweets. Then, standardization of text data was conducted through tokenization, lowercasing, punctuation removal, stop word removal, and stemming or lemmatization. Emojis and emoticons were converted into text representations using the emot library [1] Next, abbrevia-tions were expanded using a predefined dictionary, and additional punctuation marks and stopwords were removed to further refine the text data.

---
[1] https://pypi.org/project/emoji/

Table 3: Statistics of Social Support data set

| Tasks | Category | Number of samples |
|---|---|---|
| Task 1 | SS | 2236 |
|  | NSS | 7762 |
| Task 2 | Individual | 417 |
|  | Group | 1805 |
| Task 3 | Nation | 980 |
|  | Other | 512 |
|  | LGBTQ | 155 |
|  | Black people | 115 |
|  | Women | 24 |
|  | Religion | 18 |

| Comments | Task1 | Task2 | Task3 |
|---|---|---|---|
| Supporting the black community is essential for social justice | Supportive | Group | Black people |
| Women deserve equal opportunities and respect in all fields. | Supportive | Group | Women |
| oh made tear joy kids faces finally able go school precious sad video child labor glad got help angels smiling face halo hope goes well | Supportive | Group | Other |
| ronaldo football goat anything else slander wrong objectively false | Supportive | Individual |  |
| The weather today is really nice. | Non-Supportive |  |  |

Table 4: Examples for various subtasks.

### 3.3. Experiments

#### 3.3.1. Transformers

Transformers are a type of deep learning model architecture known for their effectiveness in natural language processing tasks such as translation, summarization, and text generation. They utilize self-attention mechanisms to capture long-range dependencies in text, enabling the processing of entire sentences simultaneously. The Hugging Face platform is a prominent provider of transformer-based tools and models, offering an extensive library called Transformers. This library includes pre-trained models like BERT, GPT, and T5, which can be easily fine-tuned for various NLP tasks. Hugging Face also provides user-friendly APIs and a model hub, facilitating seamless integration and deployment of transformer models in applications.

#### 3.3.2. Zeroshot

Zero-shot learning is a machine learning technique where a model is able to make predictions on new, previously unseen classes or data without having been explicitly trained on them. This is achieved by leveraging knowledge from related tasks or using semantic information about the unseen classes, allowing the model to generalize its understanding to novel situations.

Zero-shot learning enables the models to classify comments as "Supportive" or "Non-Supportive" without requiring additional training data. Specific prompts were designed to instruct the models to determine the supportiveness based solely on the given comment. Multiple prompts were used to improve reliability and the most frequent response was selected as the final prediction. This method leverages the models' ability to understand and analyze text contextually, providing accurate sentiment analysis without the need for extensive labeled datasets. The parameter details can be found in Table 3. This work utilizes models like DeBERTa and BART, including variants such as

- MoritzLaurer/deberta-v3-large-zeroshot-v2.0
- MoritzLaurer/mDeBERTa-v3-base-mnli-xnli
- MoritzLaurer/DeBERTa-v3-base-mnli-fever-anli
- sileod/deberta-v3-base-tasksource-nli
- facebook/bart-large-mnli

| Prompt | Prompt Text |
|---|---|
| 1 | Is the following comment supportive or non-supportive? We don't need any extra information, just we need the label supportive or non-supportive 'comment' |
| 2 | Determine whether this comment is supportive or non-supportive: We don't need any extra information, just we need the label supportive or non-supportive 'comment' |

Table 5: Prompts utilized for zero-shot learning to predict various types of social support

These prompts were provided to the models, and their responses were analyzed to determine the final supportiveness label for each comment.

### 3.4. Dataset Balancing

In this study, the dataset was divided into training and testing subsets, with their distributions outlined in Table 6. The BERT/DistilBERT model was utilized for classification. To balance the dataset, the k-means algorithm was applied, as it is a widely accepted and standard baseline method for clustering.

Table 6: Number of samples for Normal and Balanced datasets across different tasks and categories

| Tasks | Category | Number of Normal samples | | Number of Balanced samples | |
|---|---|---|---|---|---|
|  |  | Test | Train | Test | Train |
| Task 1 | SS | 422 | 1814 | 422 | 1814 |
|  | NSS | 1578 | 6184 | 1578 | 1814 |
| Task 2 | Individual | 87 | 336 | 87 | 336 |
|  | Group | 361 | 1452 | 361 | 336 |
| Task3 | Nation | 188 | 794 | 188 | 15 |
|  | Other | 105 | 415 | 105 | 15 |
|  | LGBTQ | 33 | 121 | 33 | 15 |
|  | Black people | 26 | 88 | 26 | 15 |
|  | Women | 7 | 17 | 7 | 15 |
|  | Religion | 4 | 15 | 4 | 15 |

## 4. Results and Analysis

#### 4.0.1. Transformers

Table 7 provides a comprehensive comparison of transformer-based models across different tasks, highlighting their performance measured by the macro F1-score.

In Task 1, models such as 'bert-base-multilingual-cased' and 'distilbert-base-uncased' demonstrated strong macro F1-scores of 0.7636 and 0.7746, respectively, indicating their effectiveness in classifying comments as supportive or non-supportive. Task 2 showcased similar trends, with models like 'bert-base-multilingual-cased' and 'roberta-base' achieving impressive macro F1-scores of 0.8146 and 0.8357, respectively, reflecting their ability to handle varying comment contexts. Meanwhile, in Task 3, 'distilbert-base-uncased' and 'roberta-base' emerged as top performers, exhibiting macro F1-scores of 0.7863 and 0.7951, respectively, underscoring their proficiency in accurately classifying comments across different supportiveness levels.

Table 7: Results for different transformers models

| Model | Tasks | Weighted scores | | | Macro scores | | | Accuracy |
|---|---|---|---|---|---|---|---|---|
| | | precision | recall | F1-score | precision | recall | F1-score | |
| bert-base-multilingual-cased | Task1 | 0.8384 | 0.8440 | 0.8396 | 0.7727 | 0.7604 | 0.7636 | 0.8440 |
| distilbert-base-uncased | | 0.8504 | 0.8568 | 0.8495 | 0.7968 | 0.7681 | 0.7746 | 0.8568 |
| google/electra-base-generator | | 0.8439 | 0.8477 | 0.8433 | 0.7804 | 0.7672 | 0.7691 | 0.8477 |
| xlnet-base-cased | | 0.8454 | 0.8489 | 0.8457 | 0.7796 | 0.7737 | 0.7740 | 0.8489 |
| albert-base-v2 | | 0.8348 | 0.8365 | 0.8316 | 0.7632 | 0.7568 | 0.7523 | 0.8365 |
| roberta-base | | 0.8502 | 0.8517 | 0.8488 | 0.7852 | 0.7809 | 0.7792 | 0.8517 |
| bert-base-multilingual-cased | Task2 | 0.8872 | 0.8890 | 0.8876 | 0.8241 | 0.8075 | 0.8146 | 0.8890 |
| distilbert-base-uncased | | 0.8978 | 0.8984 | 0.8974 | 0.8418 | 0.8241 | 0.8314 | 0.8984 |
| google/electra-base-generator | | 0.8941 | 0.8958 | 0.8945 | 0.8337 | 0.8206 | 0.8262 | 0.8958 |
| xlnet-base-cased | | 0.8939 | 0.8966 | 0.8950 | 0.8371 | 0.8166 | 0.8261 | 0.8966 |
| albert-base-v2 | | 0.8903 | 0.8931 | 0.8914 | 0.8304 | 0.8117 | 0.8205 | 0.8931 |
| roberta-base | | 0.9004 | 0.9029 | 0.9010 | 0.8495 | 0.8249 | 0.8357 | 0.9029 |
| bert-base-multilingual-cased | Task3 | 0.8619 | 0.8576 | 0.8562 | 0.7705 | 0.7616 | 0.7560 | 0.8576 |
| distilbert-base-uncased | | 0.8816 | 0.8736 | 0.8724 | 0.8351 | 0.7788 | 0.7863 | 0.8736 |
| google/electra-base-generator | | 0.8553 | 0.8538 | 0.8502 | 0.6989 | 0.7131 | 0.6953 | 0.8538 |
| xlnet-base-cased | | 0.8779 | 0.8703 | 0.8686 | 0.8279 | 0.7738 | 0.7687 | 0.8703 |
| albert-base-v2 | | 0.8677 | 0.8631 | 0.8623 | 0.8022 | 0.7590 | 0.7665 | 0.8631 |
| roberta-base | | 0.8802 | 0.8725 | 0.8729 | 0.8212 | 0.8088 | 0.7951 | 0.8725 |

### 4.0.2. Zero-shot setting

Table 8 presents the performance of various Hugging Face models across different tasks, with a focus on the macro F1-score. For Task 1, the 'MoritzLaurer/deberta-v3-large-zeroshot-v2.0' model achieved a macro F1-score of 0.57, while other models like 'mDeBERTa-v3-base-mnli-xnli' and 'DeBERTa-v3-base-mnli-fever-anli' yielded macro F1-scores ranging from 0.47 to 0.58, indicating varying degrees of effectiveness in classifying comments for supportiveness. In Task 2, the 'deberta-v3-large-zeroshot-v2.0' model demonstrated a macro F1-score of 0.39, while other models ranged from 0.40 to 0.64, suggesting differing capabilities in handling comment classification tasks. Similarly, for Task 3, models exhibited macro F1-scores ranging from 0.24 to 0.48, with the 'deberta-v3-large-zeroshot-v2.0' model achieving a score of 0.48, showcasing variations in performance across different models and tasks.

Table 8: Results for different zero-shot models

| Models | Tasks | Weighted scores | | | Macro scores | | | Accuracy |
|---|---|---|---|---|---|---|---|---|
| | | precision | recall | F1-score | precision | recall | F1-score | |
| deberta-v3-large-zeroshot-v2.0 | Task1 | 0.77 | 0.60 | 0.64 | 0.61 | 0.66 | 0.57 | 0.60 |
| mDeBERTa-v3-base-mnli-xnli | | 0.72 | 0.48 | 0.52 | 0.56 | 0.58 | 0.47 | 0.48 |
| DeBERTa-v3-base-mnli-fever-anli | | 0.70 | 0.54 | 0.58 | 0.55 | 0.57 | 0.51 | 0.54 |
| deberta-v3-base-tasksource-nli | | 0.72 | 0.51 | 0.54 | 0.56 | 0.59 | 0.49 | 0.51 |
| bart-large-mnli | | 0.72 | 0.60 | 0.63 | 0.57 | 0.60 | 0.55 | 0.60 |
| deberta-v3-large-zeroshot-v2.0 | Task2 | 0.80 | 0.39 | 0.41 | 0.58 | 0.59 | 0.39 | 0.39 |
| mDeBERTa-v3-base-mnli-xnli | | 0.76 | 0.40 | 0.43 | 0.55 | 0.57 | 0.40 | 0.40 |
| DeBERTa-v3-base-mnli-fever-anli | | 0.78 | 0.48 | 0.52 | 0.58 | 0.62 | 0.47 | 0.48 |
| deberta-v3-base-tasksource-nli | | 0.80 | 0.29 | 0.25 | 0.57 | 0.55 | 0.28 | 0.29 |
| bart-large-mnli | | 0.78 | 0.78 | 0.78 | 0.64 | 0.64 | 0.64 | 0.78 |
| deberta-v3-large-zeroshot-v2.0 | Task3 | 0.72 | 0.37 | 0.45 | 0.55 | 0.70 | 0.48 | 0.37 |
| mDeBERTa-v3-base-mnli-xnli | | 0.61 | 0.42 | 0.35 | 0.47 | 0.60 | 0.45 | 0.42 |
| DeBERTa-v3-base-mnli-fever-anli | | 0.58 | 0.31 | 0.27 | 0.38 | 0.58 | 0.34 | 0.31 |
| deberta-v3-base-tasksource-nli | | 0.58 | 0.30 | 0.24 | 0.43 | 0.59 | 0.36 | 0.30 |
| bart-large-mnli | | 0.65 | 0.60 | 0.61 | 0.46 | 0.65 | 0.50 | 0.60 |

### 4.0.3. GPT models

Table 9 presents an analysis of different models across multiple tasks. The models, including GPT3-Turbo, GPT4-Turbo, and GPT4-O, are evaluated based on precision, recall, F1-score, and accuracy for each task. For Task 1, the highest F1-score of 0.78 is achieved by GPT4-O, while the lowest F1-score of 0.62 is observed with GPT3-Turbo. In Task 2, all models show improved performance, with the highest F1-score of 0.89 attained by GPT4-O. However, in Task 3, the lowest F1-score of 0.52 is recorded for GPT4-Turbo, indicating poorer performance compared to the other models. Overall, strong performance across tasks is consistently demonstrated by GPT4-O, while varying degrees of effectiveness are exhibited by GPT4-Turbo and GPT3-Turbo depending on the task. Further insights into model performance and their respective strengths and weaknesses can be gleaned from the weighted and macro scores provided in the table.

Table 9: Results for different zero-shot models

| Model | Tasks | Weighted scores | | | Macro scores | | | Accuracy |
|---|---|---|---|---|---|---|---|---|
| | | precision | recall | F1-score | precision | recall | F1-score | |
| GPT3-Turbo | Task1 | 0.75 | 0.78 | 0.75 | 0.67 | 0.60 | 0.62 | 0.78 |
| GPT4-Turbo | | 0.77 | 0.72 | 0.74 | 0.64 | 0.68 | 0.65 | 0.72 |
| GPT4-O | | 0.78 | 0.77 | 0.78 | 0.68 | 0.69 | 0.68 | 0.77 |
| GPT3-Turbo | Task2 | 0.85 | 0.86 | 0.85 | 0.76 | 0.75 | 0.76 | 0.86 |
| GPT4-Turbo | | 0.88 | 0.89 | 0.88 | 0.83 | 0.77 | 0.80 | 0.89 |
| GPT4-O | | 0.89 | 0.88 | 0.89 | 0.80 | 0.84 | 0.82 | 0.88 |
| GPT3-Turbo | Task3 | 0.76 | 0.57 | 0.64 | 0.67 | 0.72 | 0.62 | 0.57 |
| GPT4-Turbo | | 0.73 | 0.44 | 0.52 | 0.65 | 0.70 | 0.58 | 0.44 |
| GPT4-O | | 0.83 | 0.60 | 0.66 | 0.70 | 0.77 | 0.63 | 0.60 |

### 4.1. Balanced data sets

To balance the dataset, the k-means algorithm was employed, as it is a widely recognized and standard baseline method for clustering. The bar chart illustrates the F1-scores for three tasks when using normal and balanced datasets. For Task1 and Task2, the normal dataset outperformed the balanced version, achieving F1-scores of 0.80 and 0.86, respectively, compared to 0.75 in both cases for the balanced dataset. However, Task3, which involves six labels with a highly skewed distribution, experienced a significant performance drop. The F1-score fell sharply from 0.67 (normal) to just 0.11 (balanced).

This drastic decline in Task3 can be attributed to two key factors. First, the severe imbalance in label distribution resulted in some classes having far fewer examples than others. When k-means clustering was applied, it attempted to balance the dataset by oversampling minority classes and undersampling majority classes. However, this process led to the removal of a substantial amount of valuable data, including critical information from the majority classes. You can find the results in Tables 10 and 11

Table 10: Performance metrics of the roberta-base model on a normal dataset

| Task | Model | Weighted Scores | | | Macro Scores | | | Accuracy |
|---|---|---|---|---|---|---|---|---|
| | | precision | recall | F1-score | precision | recall | F1-score | |
| Task1 | roberta-base | 0.87 | 0.86 | 0.87 | 0.79 | 0.82 | 0.80 | 0.86 |
| Task2 | | 0.91 | 0.91 | 0.91 | 0.86 | 0.86 | 0.86 | 0.91 |
| Task3 | | 0.83 | 0.83 | 0.83 | 0.72 | 0.66 | 0.67 | 0.83 |

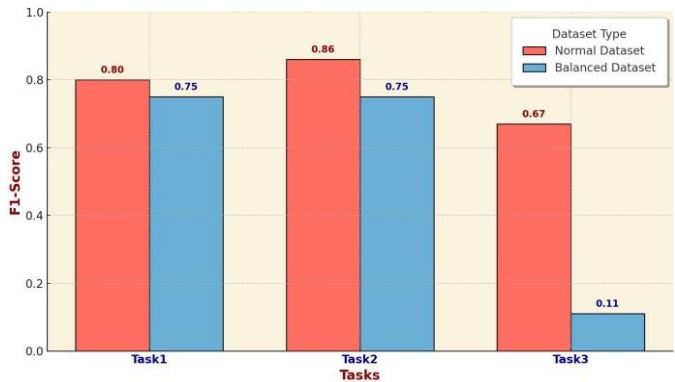

Figure 1: F1-Score comparison for Normal vs Balanced Datasets

Table 11: Performance metrics of the roberta-base model on a balanced dataset

| Task | Model | Weighted Scores | | | Macro Scores | | | Accuracy |
|---|---|---|---|---|---|---|---|---|
| | | precision | recall | F1-score | precision | recall | F1-score | |
| Task1 | roberta-base | 0.85 | 0.81 | 0.82 | 0.73 | 0.80 | 0.75 | 0.81 |
| Task2 | | 0.87 | 0.81 | 0.83 | 0.73 | 0.82 | 0.75 | 0.81 |
| Task3 | | 0.27 | 0.52 | 0.35 | 0.09 | 0.17 | 0.11 | 0.52 |

## 5. Error analysis

The confusion matrix presented shows the performance of a classification model distinguishing between "Supportive" and "Non-Supportive" instances. The overall accuracy of the model is 84.21%, with a misclassification rate of 15.79%. The model performs well in identifying "Non-Supportive" instances, achieving an 89.00% true positive rate. However, it struggles more with "Supportive" instances, with only a 67.58% true positive rate and a relatively high false negative rate of 32.42%. This indicates that the model is more prone to incorrectly labeling "Supportive" instances as "Non-Supportive" compared to the reverse. Improving the model's sensitivity to "Supportive" instances while maintaining its strong performance on "Non-Supportive" instances would enhance overall accuracy.

The confusion matrix indicates that the classification model has an overall accuracy of 89.98%, with a misclassification rate of 10.02%. For the "Group" class, the model correctly identified 94.32% of instances, but incorrectly classified 5.68% as "Individual". For the "Individual" class, 71.39% were correctly identified, while 28.61% were misclassified as "Group". This disparity suggests the model is more proficient at identifying "Group" instances compared to "Individual" instances, which are misclassified at a higher rate. Improving the classification of "Individual" instances could enhance overall performance.

The confusion matrix for this multi-class classification problem shows varying levels of accuracy across different classes. The model performs exceptionally well in classifying "Nation" (92.97%) and "LGBTQ" (94.81%) instances. However, it struggles with "Religion," where it correctly classifies only 47.37% of instances, misclassifying a significant portion as "Nation." "Other" has a moderate classification accuracy of 75.19%, with notable misclassifications across other classes, particularly "Nation" (16.35%). The "Black Community" and

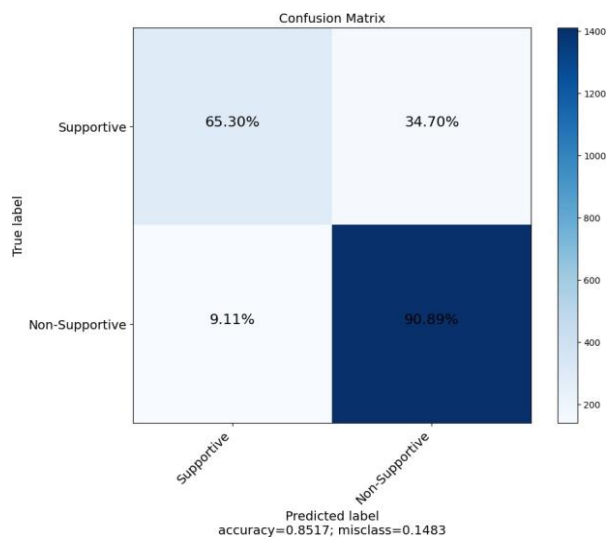

Figure 2: Confusion matrix depicting supportive and non-supportive classifications

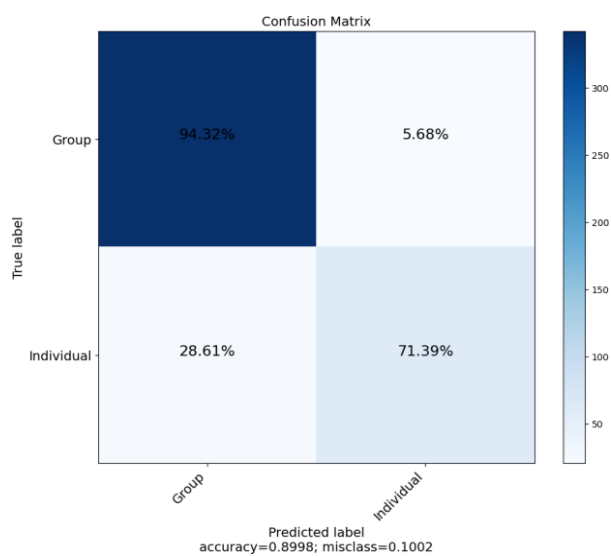

Figure 3: Confusion matrix depicting Group and Individual classifications

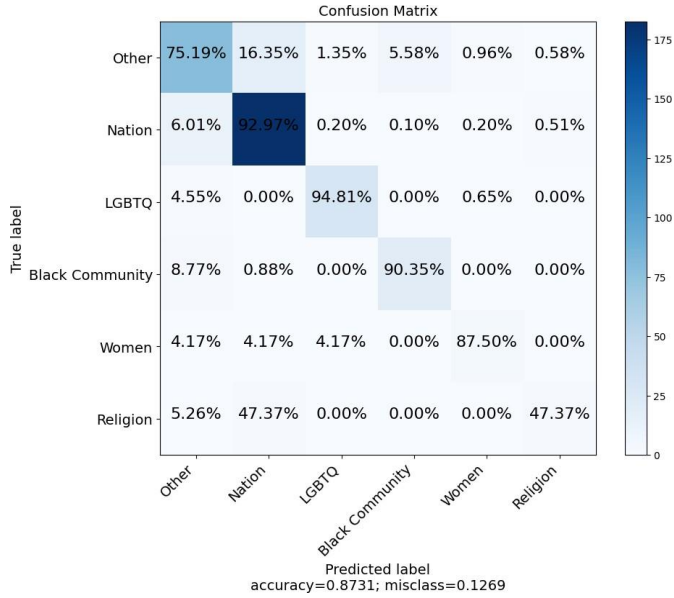

Figure 4: Confusion matrix illustrating various classifications within the Group category

"Women" classes have reasonable accuracies of 90.35% and 87.50%, respectively, but still show some confusion, especially "Women" being equally misclassified into several categories. Overall, the model achieves an accuracy of 87.31% and a misclassification rate of 12.69%, indicating that while the model generally performs well, there are specific areas, particularly "Religion" and to some extent "Other," where improvements are needed. A graphical representation of the confusion matrix is presented in Table 12.

Table 12: Classwise scores for best performing model for each subtask

| Tasks | Model | Label | precision | recall | F1-score |
|---|---|---|---|---|---|
| SubTask 1 | roberta-base | SS | 0.6675 | 0.6529 | 0.6531 |
| | | NSS | 0.9028 | 0.9089 | 0.9052 |
| SubTask 2 | | Individual | 0.7469 | 0.7139 | 0.7329 |
| | | Group | 0.9341 | 0.9432 | 0.9385 |
| SubTask 3 | | Other | 0.8349 | 0.7519 | 0.7894 |
| | | Nation | 0.9057 | 0.9297 | 0.9173 |
| | | LGBTQ | 0.9397 | 0.9477 | 0.9425 |
| | | Black Community | 0.8517 | 0.9020 | 0.8519 |
| | | Women | 0.8111 | 0.8800 | 0.8040 |
| | | Religion | 0.5733 | 0.5000 | 0.4776 |

## 6. Discussion

The current study reveals that users on YouTube tend to express more support for groups than for individuals, with significant support for different nations without heavy religious influence. Recent data indicates a focus on nations and communities like LGBTQ+ individuals and Black people, highlighting a social emphasis on national and community identities over religious considerations. This study serves as a foundational step in introducing the task of social support detection aimed at fostering support and positivity as an alternative to merely filtering out hate speech.

## 7. Conclusions and future work

In this study, we utilized a novel dataset focused on social support, categorized into individual and group levels. The group category was further subdivided into various segments, including Nation, Other, LGBTQ, Black people, Women, and Religion. To analyze this data, we employed several models, starting with zero-shot learning using large language models (LLMs) such as GPT-3, GPT-4, and GPT-4-turbo.

For classification, we also utilized a range of models available on Hugging Face, the results of which are detailed in Table 4. Our approach yielded significant improvements in macro F1 scores, with an increase of 7% in Task 2 and 8% in Task 3. Notably, the transformers model, roberta-base, consistently demonstrated superior performance across all three tasks, outperforming all other models we tested.

Looking ahead, our future work will focus on integrating different LLMs, enhancing prompt engineering, and exploring few-shot learning techniques to further improve classification accuracy.

## 8. Limitation

Despite the promising results, our study has several limitations. Firstly, the novel dataset on social support, while comprehensive, may have inherent biases due to its specific categorization into individual and group levels and further subdivision into segments such as Nation, Other, LGBTQ, Black people, Women, and Religion. These categorizations might not fully capture the complexity and nuances of social support across different contexts. Additionally, while we employed advanced models including zero-shot learning with GPT-3, GPT-4, and GPT-4-turbo, the generalizability of these models is limited by the quality and diversity of the training data. Furthermore, the superior performance of the roberta-base transformer model across all tasks suggests a potential over-reliance on this specific architecture, which may not necessarily translate to other datasets or real-world applications. Lastly, our current approach does not fully leverage few-shot learning, which could potentially enhance model performance in low-resource scenarios.

**Declaration of Interest Statement**

*Funding*

The work was done with partial support from the Mexican Government through the grant A1-S-47854 of CONACYT, Mexico, grants 20241816, 20241819, and 20240951 of the Secretaría de Investigación y Posgrado of the Instituto Politécnico Nacional, Mexico. The authors thank the CONACYT for the computing resources brought to them through the Plataforma de Aprendizaje Profundo para Tecnologías del Lenguaje of the Laboratorio de Supercómputo of the INAOE, Mexico and acknowledge the support of Microsoft through the Microsoft Latin America PhD Award.